\documentclass[10pt,twocolumn,letterpaper]{article}

\usepackage[accsupp]{axessibility}
\usepackage{iccv}

\usepackage{changepage,threeparttable} 
\usepackage{multirow} 
\usepackage{changepage} 

\usepackage{amsmath}
\usepackage{amsfonts}
\usepackage{amssymb}
\usepackage{amsthm}
\usepackage{bm}
\usepackage{bbm}
\usepackage{mathtools}
\usepackage{enumitem}
\usepackage{thmtools,thm-restate}
\usepackage[ruled,vlined,linesnumbered]{algorithm2e}
\SetAlFnt{\small}
\usepackage{color}
\usepackage[dvipsnames]{xcolor}
\usepackage{graphicx}
\usepackage{comment}

\theoremstyle{plain}

\theoremstyle{definition}

\theoremstyle{remark}

\def\btheta{{\bm\theta}}

\def\DD{\mathcal{D}}

\def\NN{\mathcal{N}}
\def\PP{\mathcal{P}}
\def\SX{\mathcal{S}}\def\TT{\mathcal{T}}

\def\lb{\mathbf{l}}

\def\pb{\mathbf{p}}

\def\vb{\mathbf{v}}

\def\Rbb{\mathbb{R}}

\def\R{\Rbb}

\def\*{\star}

\usepackage{etex,etoolbox}
\usepackage{environ}

\makeatletter
\providecommand{\@fourthoffour}[4]{#4}
\newcommand\fixstatement[2][\proofname\space of]{%
	\ifcsname thmt@original@#2\endcsname
	\AtEndEnvironment{#2}{%
		\xdef\pat@label{\expandafter\expandafter\expandafter
			\@fourthoffour\csname thmt@original@#2\endcsname\space\@currentlabel}%
		\xdef\pat@proofof{\@nameuse{pat@proofof@#2}}%
	}%
	\else
	\AtEndEnvironment{#2}{%
		\xdef\pat@label{\expandafter\expandafter\expandafter
			\@fourthoffour\csname #1\endcsname\space\@currentlabel}%
		\xdef\pat@proofof{\@nameuse{pat@proofof@#2}}%
	}%
	\fi
	\@namedef{pat@proofof@#2}{#1}%
}

\newcounter{proofcount}

\NewEnviron{proofatend}{%
	\edef\next{%
		\noexpand\begin{proof}[\pat@proofof\space\pat@label]%
			\unexpanded\expandafter{\BODY}}%
		\global\toks\numexpr\prooftoks+\value{proofcount}\relax=\expandafter{\next\end{proof}}
	\stepcounter{proofcount}}

\def\printproofs{%
	\count@=\z@
	\loop
	\the\toks\numexpr\prooftoks+\count@\relax
	\ifnum\count@<\value{proofcount}%
	\advance\count@\@ne
	\repeat}
\makeatother

\fixstatement{lemma}
\fixstatement{theorem}
\fixstatement{proposition}
\fixstatement{corollary}
\usepackage{times}
\usepackage{epsfig}
\usepackage{graphicx}
\usepackage{amsmath}
\usepackage{amssymb}

\usepackage{booktabs}
\usepackage{pifont}
\usepackage{xcolor}
\definecolor{blackpink}{rgb}{1.0, 0.61, 1.0}
\newcommand{\drule}{\specialrule{0.2pt}{1pt}{1pt} \specialrule{0.2pt}{0pt}{\belowrulesep}}

\usepackage{hyperref}
\hypersetup{pagebackref=true,breaklinks=true,letterpaper=true,colorlinks,bookmarks=false}
\usepackage{cleveref}
\usepackage{url}
\iccvfinalcopy 

\def\iccvPaperID{1257} 

\ificcvfinal\fi

\begin{document}

\title{LiDAR-UDA: Self-ensembling Through Time for Unsupervised LiDAR Domain Adaptation}

\author{First Author\\
Institution1\\
Institution1 address\\
{\tt\small firstauthor@i1.org}
\andLiDAR Domain Adaptation
Second Author\\
Institution2\\
First line of institution2 address\\
{\tt\small secondauthor@i2.org}
}
\author{Amirreza Shaban\thanks{Equal Contribution.}\quad
 JoonHo Lee$^*$\quad
 Sanghun Jung$^*$\quad
 Xiangyun Meng\quad
 Byron Boots\quad
 \and
 University of Washington
 }
 
\maketitle
\ificcvfinal\thispagestyle{empty}\fi

\begin{abstract}
We introduce LiDAR-UDA, a novel two-stage self-training-based Unsupervised Domain Adaptation (UDA) method for LiDAR segmentation. Existing self-training methods use a model trained on labeled source data to generate pseudo labels for target data and refine the predictions via fine-tuning the network on the pseudo labels. These methods suffer from domain shifts caused by different LiDAR sensor configurations in the source and target domains. We propose two techniques to reduce sensor discrepancy and improve pseudo label quality: 1) LiDAR beam subsampling, which simulates different LiDAR scanning patterns by randomly dropping beams; 2) cross-frame ensembling, which exploits temporal consistency of consecutive frames to generate more reliable pseudo labels. Our method is simple, generalizable, and does not incur any extra inference cost. We evaluate our method on several public LiDAR datasets and show that it outperforms the state-of-the-art methods by more than $3.9\%$ mIoU on average for all scenarios. Code will be available at \href{https://github.com/JHLee0513/LiDARUDA}{https://github.com/JHLee0513/LiDARUDA}.

\end{abstract}

\section{Introduction}
Modern approaches to perception for robotics and autonomous driving rely on supervised LiDAR segmentation methods that can accurately identify objects and scenes from 3D point clouds. These methods have advanced significantly thanks to large public datasets~\cite{behley2019iccv,nuscenes2019,Sun_2020_CVPR,pan2020semanticposs} that enable the development of efficient and powerful deep neural networks~\cite{zhu2020cylindrical,Cheng2021AF2S3NetAF,yan20222dpass}, and they have inspired applications in other domains such as off-road navigation~\cite{Maturana2017RealTimeSM,shaban2021semantic}, locomotion navigation~\cite{Frey_2022}, and construction site mapping~\cite{guan2021tns}. 
However, supervised LiDAR segmentation often struggles to adapt to new domains (\textit{i.e.,} domains that the model is not trained on) due to distributional shifts~\cite{datasetshift} between source and target datasets. LiDAR perception poses a unique challenge in this regard because different sensor configurations (\textit{e.g.,} beam patterns, reflectivity estimates, mounting position, etc.) introduce significant distributional shifts~\cite{triess2021iv}. To mitigate such concerns, several Unsupervised Domain Adaptation (UDA) approaches~\cite{langer2020domain, jiang2020lidarnet, comp_label, graph_matching} have been proposed, transferring the knowledge of a model trained on one domain to another without requiring additional labels.  UDA plays an essential role in LiDAR segmentation since it relieves the necessity of an expensive and labor-intensive labeling process.

Domain adaptation methods based on \emph{self-training}, which work by iteratively generating pseudo labels on target data and retraining the model with these labels, have achieved great success in reducing covariate shift in image-based semantic segmentation tasks~\cite{IAST,Zou2018UnsupervisedDA,french2018selfensembling,araslanov2021self}. These self-training methods operate under the assumption that a model trained on source data yields mostly accurate predictions on at least a subset of the target dataset, enabling the model to adapt and refine its predictions iteratively through fine-tuning over the pseudo labels. 
However, in the case of LiDAR segmentation, the beam pattern gap between different LiDAR sensors hampers the source model from predicting reasonably good pseudo labels in the target domain for initializing the self-training approach.

To overcome this gap between the source and target datasets, we propose a simple yet effective structured point cloud subsampling method that simulates different LiDAR beam patterns. Specifically, we randomly subsample rows in the range image~\cite{rangenet++} of a high-beam LiDAR sensor to simulate low-beam LiDAR sensors. 
Additionally, we propose \textit{cross-frame ensembling}, a temporal ensembling module, to ensure consistency of pseudo labels across LiDAR scans within each sequence. Cross-frame ensembling aggregates predictions from multiple scans and uses nearest neighbors to refine the pseudo labels. While we could simply calculate the average with uniform weights, this method ignores the temporal (i.e., time from the reference scan) and spatial variations (i.e., distance to sensor origin for each scan) of points captured by the LiDAR sensor when aggregating multiple scans. We address this issue by training a Learned Aggregation Model (LAM) that resembles graph convolution~\cite{thomas2019KPConv, wang2018deep}. LAM learns how to aggregate pseudo labels within a sequence and weigh labels for each point differently according to its importance.
Adopting this approach eliminates the need for ad-hoc approaches to deal with special cases such as moving objects~\cite{comp_label, langer2020domain}.

We show that the combination of proposed modules achieves state-of-the-art performance in domain adaptation scenarios for urban and off-road driving. Moreover, our framework is applicable to off-the-shelf LiDAR segmentation networks since it does not require any architectural modifications or impose additional computational costs during inference. In contrast to previous work~\cite{langer2020domain, comp_label} that uses aggregated LiDAR scans within a sequence as a dense and sensor-agnostic representation for the segmentation network, our approach maintains the sparsity of the point cloud during the network forward pass. This characteristic enables us to use state-of-the-art network architectures that favor sparse convolutions for efficient LiDAR segmentation.

\section{Related Work}

\textbf{LiDAR Semantic Segmentation} LiDAR semantic segmentation is a fundamental capability for scene understanding in autonomous driving and robotics. In recent years, deep learning methods have achieved remarkable results on LiDAR segmentation, thanks to several large-scale datasets and benchmarks, such as nuScenes \cite{nuscenes2019}, SemanticKITTI \cite{behley2019iccv}, and SemanticPOSS \cite{pan2020semanticposs}. Approaches to LiDAR segmentation can be broadly classified into point-based \cite{rangenet++, hu2020randla}, image-based \cite{cortinhal2020salsanext, squeezesegv2}, sparse voxel-based \cite{zhu2020cylindrical}, and hybrid \cite{spvnas} categories. Despite remarkable progress, existing methods still face challenges in generalizing to different datasets due to two factors: 1) different datasets have different semantic classes and geometric feature distributions, depending on the environments where they were collected. 2) LiDAR sensors have different mounting positions and produce different beam patterns. Therefore, models trained on one dataset may not perform well on another dataset~\cite{comp_label, jiang2020lidarnet,langer2020domain}. This limitation restricts the practical applicability of LiDAR-based segmentation methods because labeling LiDAR points is costly and time-consuming.

\textbf{Domain Adaptation for LiDAR} There has been an increasing interest in developing domain adaptation techniques to improve the generalization ability of LiDAR perception models across different LiDAR sensors and environments. Domain adaptation methods for LiDAR can be grouped into three categories, which we describe here. 1) \emph{Learning domain-invariant representation}. These methods transform the source and target domain point clouds into a common representation that is independent of sensor characteristics.  The common representation can be a 3D mesh~\cite{langer2020domain, comp_label} or a bird’s eye view projection~\cite{piewak2019analyzing}. Notably, Complete \& Label~\cite{comp_label} learns to complete 3D surfaces from sparse LiDAR scans using a sensor-specific network, and then applies a segmentation network on the completed surfaces. However, this method requires a simplified segmentation network to handle the dense point clouds and also needs to remove moving objects from the common representation using heuristic methods~\cite{comp_label} or manual annotations~\cite{langer2020domain}. We use a data-driven approach to decide how different semantic classes should be aggregated. 2) \emph{Learning domain-invariant features.} These methods align or adapt the feature representations of source and target domains using various techniques, such as feature alignment \cite{squeezesegv2,morerio2017minimal}, adversarial training \cite{jiang2020lidarnet}, multi-task learning \cite{rist2019cross}, and graph matching \cite{graph_matching}. These methods do not modify the input point clouds but learn to extract features that are robust to domain variations. 3) \emph{Domain transfer}. These methods explicitly model the difference between source and target domains and apply it to transfer one domain to another. For example, some methods learn a noise model from real data and add it to synthetic data to make it more realistic \cite{nuscenes2019}. Our method applies LiDAR beam subsampling to reduce the domain gap without needing heuristics on which rows to drop and instead uses a random selection scheme.

\textbf{Self-training} We leverage self-training, a semi-supervised learning technique~\cite{laine2016temporal, tarvainen2017mean} that has been successfully applied for unsupervised domain adaptation in the image domain~\cite{araslanov2021self, zou2019confidence, li2022class}, but has not been extensively explored for LiDAR domain adaptation. Self-training, also known as teacher-student training or self-ensembling, iteratively trains a model on a mixture of labeled source data and pseudo labeled target data, where the pseudo labels are generated by the model itself on unlabeled target data. However, since the pseudo labels may be noisy or inaccurate, self-training often requires some regularization strategies to improve their quality and reliability, such as class balancing~\cite{zou2019confidence}, adversarial pre-training~\cite{xia2021adaptive}, and uncertainty estimation~\cite{zheng2021rectifying}. In our approach, we adopt a teacher-student paradigm and use a data-driven aggregation scheme that selectively aggregates the pseudo labels within each sequence as a regularizer. This further enhances the performance of our model on the target domain.

\section{Method}

\subsection{Definitions and Framework Overview}
We consider the problem of point cloud semantic segmentation in a domain adaptation setting. Let $\SX$ and $\TT$ denote the source and target datasets, respectively. Each element in the source dataset consists of a tuple $(\PP, \lb)$, where $\PP \in \mathbb{R}^{P \times 3}$ represents a 3D point cloud and $\lb \in \{0,1\}^{P\times K}$ denotes the corresponding one hot semantic labels with $K$ classes. In contrast to the source dataset, the target dataset only contains unlabeled point clouds. We address the closed-set adaptation problem~\cite{triess2021iv}, where both the source and target domains share the same semantic classes. Our objective is to train a model on the labeled source dataset and unlabeled target point clouds, and then evaluate it on a held-out target set using ground-truth annotations. Our approach employs a two-stage self-ensembling strategy to learn a performant segmentation model for the target dataset. A LiDAR segmentation model $F_\theta: \mathbb{R}^{P \times 3} \to \mathbb{R}^{P\times K}$ is first trained on the labeled source dataset, and then adapted to the target dataset. The source model training and domain adaption stages are detailed next.
\begin{figure}[t!]
  \centering
  \includegraphics[width=\linewidth]{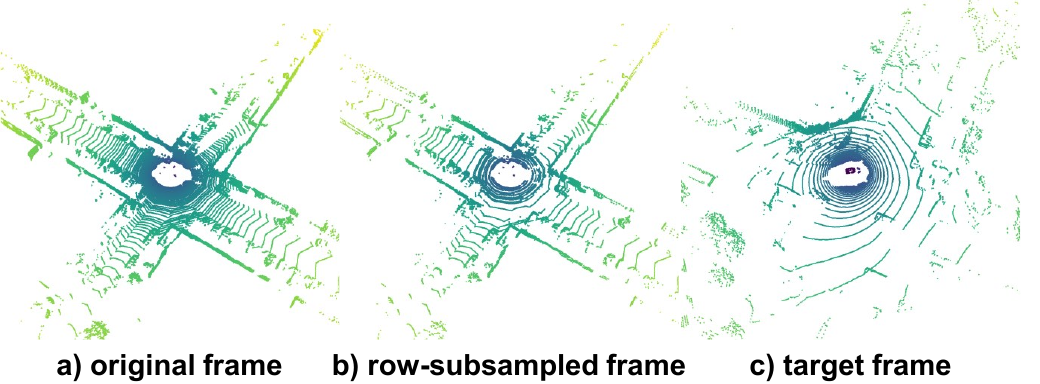}
    \caption{Comparison of a) original SemanticKITTI LiDAR scan with $64$ LiDAR beams, b) row-subsampled by randomly dropping range image rows with probability $0.5$, and c) nuScenes LiDAR scan with $32$ beams. The figure shows that subsampling rows effectively simulates LiDAR scans with fewer laser beams.}
    \label{fig:pc_subsampling}
\end{figure}
\subsection{Source Model Training}\label{sec:source_training}
We train the source model using standard supervised learning, which enables our framework to be applied to any generic LiDAR segmentation model. To facilitate generalization to target domains with fewer LiDAR beams than the source domain, we apply \emph{structured point cloud subsampling} along with conventional data augmentations during training. Specifically, we subsample the point cloud on the \emph{range image}, which is created by spherical mapping of a LiDAR scan into a 2D image~\cite{rangenet++}. The image is represented as $I_\PP \in \mathbb{R}^{H \times W \times 3}$, where $H$ and $W$ are the height and width of the projected image. The mapping from the 3D point $\pb = (x, y, z)$ to the image coordinate $(u, v)$ is defined as
\begin{equation*}
\left(\begin{array}{l}
u \\
v
\end{array}\right)=\left(\begin{array}{c}
\frac{1}{2}\left[1-\arctan (y, x) \pi^{-1}\right] W \\
{\left[1-\left(\arcsin \left(z/||\pb||_2\right)+f_{\text {down}}\right) f^{-1}\right] H}
\end{array}\right),
\end{equation*}
where $f_{\text {down}}$ and $f$ denote lower and vertical LiDAR field-of-view, respectively. 
Then, we randomly drop all the points on a horizontal line with probability $1 - \min(1, r)$, where $r=n_\text{target}/n_\text{source}$ and $n_\text{target},n_\text{source}$ are the number of laser beams in the target and source datasets, respectively. In cases where the target LiDAR has more beams than the source dataset, we do not apply subsampling to the source dataset. Instead, we address the domain gap by subsampling the target point cloud during the domain adaptation stage in \Cref{sec:within}. As demonstrated in \Cref{fig:pc_subsampling}, row subsampling effectively simulates a LiDAR scan with fewer laser beam patterns. We further elaborate on the effectiveness of this data augmentation in \Cref{subsec:ablation_study}.

\subsection{Target Domain Adaptation}\label{sec:DA}
The domain adaptation stage is an iterative process where each iteration involves generating pseudo labels using a teacher model (\textit{i.e.,} label generation step), and training a student network with given pseudo labels (\textit{i.e.,} training step). The adaptation allows for multiple iterations, where any additional iterations after the initial source-to-target adaptation may be viewed as further refinements within the target domain.

\begin{figure}[t!]
  \centering
  \includegraphics[width=\linewidth]{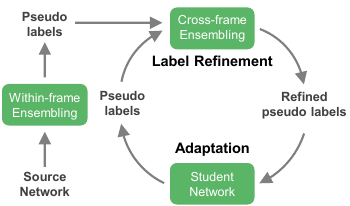}
    \caption{Overview of the domain adaptation process. We first apply within-frame ensembling with the source model (source network) $F_\theta(\cdot)$ to generate the pseudo labels. Subsequently, we apply cross-frame ensembling with the LAM module $g_\omega(\cdot)$ to refine the initially generated pseudo labels.
    Then, we adapt the student network to the target domain by training it with the refined pseudo labels for a certain number of epochs, and finally, re-generate the pseudo labels from the trained student network. The cross-frame ensembling and adaptation steps are iterated multiple times. 
    }
    
    \vspace{-0.3cm}
    \label{fig:method_overview}
\end{figure}

\Cref{fig:method_overview} summarizes the domain adaptation stage. We employ within-frame and cross-frame ensembling techniques to enhance the quality of the pseudo labels and improve the training of the student model. In the initial source-to-target adaptation iteration, we employ within-frame subsampling (\Cref{sec:within}) to reduce high-beam target beams to match the low-beam source model. We further enhance the pseudo labels by aggregating predictions within each sequence using cross-frame ensembling described in \Cref{sec:across}. 

The student network is randomly initialized and trained with aggressive data augmentation to enforce consistent predictions across different augmentations for domain adaptation, following a common practice in previous work~\cite{araslanov2021self}. While recent literature~\cite{araslanov2021self, momentumContDA} utilizes a momentum network as a teacher, we adopt a fixed teacher model that allows us to pre-compute the pseudo labels at the beginning of each domain adaptation iteration and reduce the training time significantly.

\begin{figure*}[t!]
  \centering
  \includegraphics[width=0.95\linewidth]{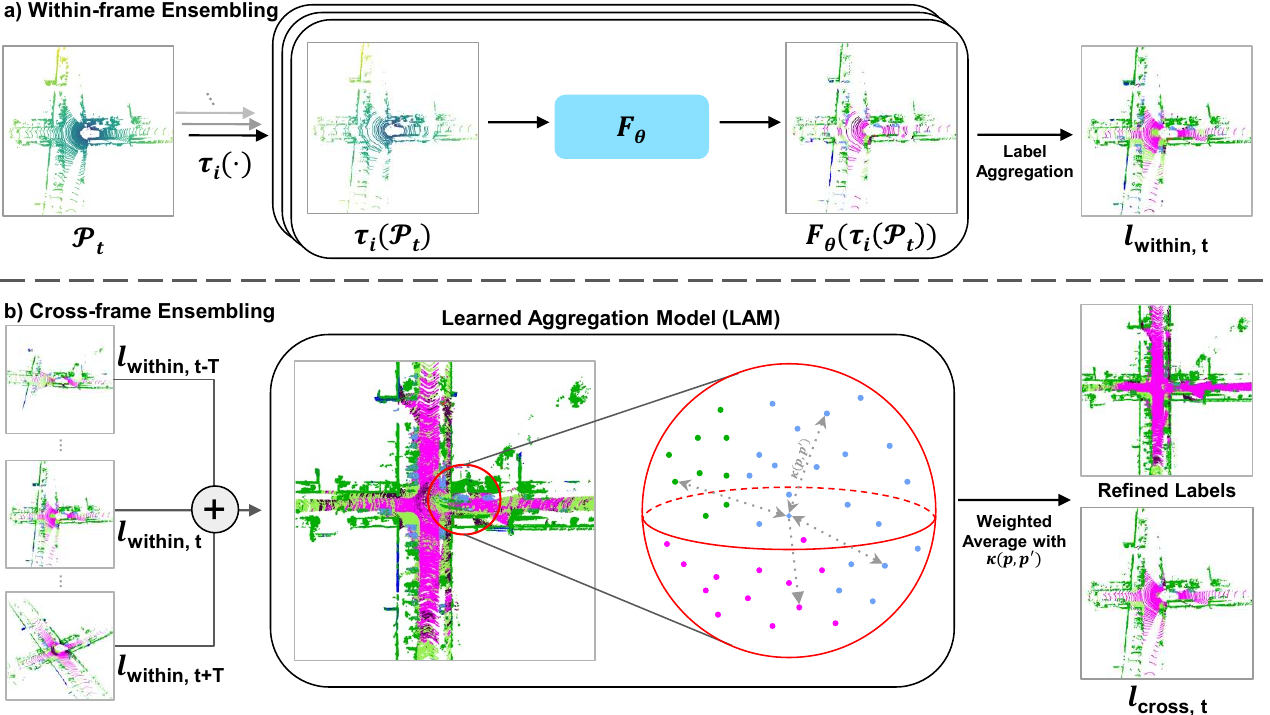}
    \caption{Illustration of our within-frame and cross-frame ensembling modules. All the predictions in the figure are obtained from our nuScenes to SemanticKITTI experiment. a) Within-frame ensembling: we randomly select horizontal rows with a probability of $1 - \min(1, 1/r)$ and drop all the points in the rows to simulate the different beam patterns of the target domain. To obtain more robust predictions, we apply this subsampling several times and average the prediction.
    b) Cross-frame ensembling: with the obtained predictions from step a), we temporally aggregate the point clouds and their predictions. Afterward, we calculate the nearest neighboring points within $\epsilon$-ball and predict their summing weight using LAM. Finally, we obtain the refined pseudo labels by weight averaging the pseudo labels of the neighboring points.
    }
    \label{fig:within_across_ensembling}
\end{figure*}
\subsubsection{Within-frame Ensembling}\label{sec:within}
We use within-frame ensembling when we have more beams in the target LiDAR than in the source LiDAR. As shown in \Cref{fig:within_across_ensembling}-a, this approach works as follows: First, we create a batch of randomly subsampled point clouds from an input LiDAR scan. Then, we use the source model to predict labels for each subsampled point cloud. Finally, we average the predictions across all subsampled point clouds to get the final prediction.

Let $\PP$ be an input point cloud. We generate a set of subsampled point clouds $\mathbb{T}(\PP) = \{\tau_i(\PP)\}_{i=1}^{N_{s}}$, where $N_{s}$ is the number of trials and $\tau(\cdot)$ is a subsampling operation. We set $\tau_1(\cdot)$ as an identity mapping to keep the original input point cloud and follow a similar approach as in \Cref{sec:source_training}, but we use the source-to-target ratio ($1/r$) to drop points within a row of the LiDAR image with a probability of $1 - \min(1, 1/r)$. Then, we obtain a set of predictions from the pretrained network $F_\btheta(\cdot)$ by computing $\mathbb{P}(\PP) = \{F_\btheta(\tau_i(\PP))\}_{i=1}^{N_{s}}$. For each point in the original point cloud $\pb \in \PP$, we compute its final prediction by averaging all the predictions associated with $\pb$ within the augmented point clouds $\mathbb{P}(\PP)$. As the original point cloud is always included, every point appears at least once during aggregation.

\subsubsection{Cross-frame Ensembling}\label{sec:across}

Our cross-frame ensembling is illustrated in \Cref{fig:within_across_ensembling}b.
To further refine the pseudo labels of individual scans, we utilize predictions on scans from both previous and subsequent timestamps. Given an input query scan $\PP_t$ at time $t$, we aggregate scans from the past and future into a dense point cloud $\DD = \bigcup_{i=t-T}^{t+T} A_i(\PP_i)$, where $A_i$ represents the transformation from index $i$ to $t$, and $T$ controls the number of aggregated frames.
Let $\NN(\pb) \subset \DD$ denote the set of points that fall in the vicinity of $\pb\in\PP_t$. We compute the enhanced class probability vector $\tilde{\vb}(\pb)$ as
\begin{equation}\label{eq:knnpp}
    \tilde{\vb}(\pb) = \frac{1}{Z(\pb)}\sum_{\pb' \in \NN(\pb)} \kappa(\pb, \pb') \vb(\pb'),
\end{equation}
where $\vb(\pb')$ is the single scan pseudo label of $\pb'$, $\kappa: \R^3\times\R^3\to\R^+$ is a positive scoring function, and $Z$ is a normalizer that ensures $\tilde{\vb}(\pb)$ remains a probability vector, i.e.,
\begin{equation}\label{eq:knnpp_norm}
    Z(\pb) = \sum_{\pb' \in \NN(\pb)} \kappa(\pb, \pb').
\end{equation}

In our experiments, we compute $\NN(\pb)$ by finding the $k$-nearest neighbors and then selecting the subset that lies within an $\epsilon$-ball (i.e., a ball with radius $\epsilon$) centered on $\pb$. This approach guarantees that the point count remains under a specific threshold, and all points are within the $\epsilon$-ball surrounding $\pb$.

We can set $\kappa(\pb, \pb')$ to a constant value to obtain the standard KNN algorithm, but this method is more suitable for static objects. Assigning the same weight to all the points in $\NN(\pb)$ overlooks their semantic classes, their distances from the query point $\pb$, and the fact that these points are captured at different times and distances from the LiDAR sensor.

We further improve the quality of refined labels by learning an attention model for label aggregation, which we refer to as the Learned Aggregation Model (LAM). To account for the various sources of aggregation error, LAM considers not only the Euclidean distance between input points $\pb$ and $\pb^\prime$, but also their single scan pseudo labels $\vb(\pb)$ and $\vb(\pb^\prime)$, the temporal offset between $\pb$ and $\pb^\prime$, and the distance of $\pb^\prime$ to its sensor origin. For ease of notation, we use $\Phi(\pb, \pb^\prime)\in\mathbb{R}^D$ to denote the feature vector that concatenates each of these factors. Then, we use a fully connected network $g_\omega:\R^D\to\R$ to predict an attention score for each feature vector. Finally, enhanced pseudo labels are obtained via attention by setting $\kappa(\pb, \pb^\prime) = \exp(g_{\omega}(\Phi(\pb, \pb^\prime)))$ in \Cref{eq:knnpp}.

\begin{figure}[t!]
  \centering
  \includegraphics[width=0.9\linewidth]{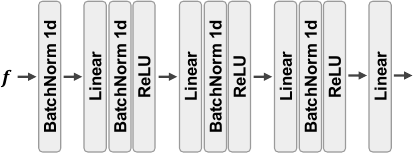}
    \caption{Architecture of $g_\omega(\cdot)$ our Learned Aggregation Model (LAM) module.
    Note that $\mathbf{f}$ denotes $\mathbf{f} = \mathbf{\Phi}(\pb, \pb^\prime)$.
    We first apply the batch normalization layer to effectively address the differences in statistics of source and target domains.
    }
    \label{fig:LAM_architecture}
\end{figure}
To train LAM, we use the source model predictions and aforementioned features as inputs, while supervision is provided through the source dataset ground-truth labels. We first compute the single-scan pseudo label $\vb(\pb)$ for each point  within the source dataset using the source model network $F_\theta(\cdot)$. Next, we construct the dense point cloud by aggregating scans and pre-compute the nearest neighbors $\NN(\pb)$ for all the points to speed up the training. During training, we estimate the enhanced labels using \Cref{eq:knnpp} and use a combination of multi-class cross-entropy and Lov{\'a}sz-Softmax loss~\cite{berman2018lovasz} to update model parameters $\omega$.

The LAM model $g_\omega$, shown in \Cref{fig:LAM_architecture}, consists of an input standardization layer and $3$ fully connected hidden layers each followed by batch-norm and ReLU layers. The collected statistics (i.e. mean and variance) for the standardization layer captured by LAM are fit to the source domain and hence are not optimal for adaptation. Therefore, we \textit{modulate} the statistics by updating the layer with statistics acquired from the target domain. Since the statistics are collected from input point features including semantic pseudo labels, it does not involve acquiring any privileged information from the target domain. We find that updating the statistics in the first layer serves a similar purpose to the Adaptive Batch Normalization (ABN) domain adaption method~\cite{li2016revisiting} that adapts the batch-norm statistics across the network.

\section{Experiments}
\subsection{Experimental Setup}\label{subsec:experimental_setup}
We compare our method against prior domain adaptation methods on publicly available LiDAR segmentation datasets using the mean Intersection over Union (mIoU) metric. In particular, our main experiment is split into two tracks: 1) between SemanticKITTI~\cite{behley2019iccv} and nuScenes~\cite{nuscenes2019}; 2) between SemanticKITTI~\cite{behley2019iccv}, SemanticPOSS~\cite{pan2020semanticposs}, and SemanticUSL~\cite{jiang2020lidarnet}. These tracks demonstrate that our method is superior to prior work in the presence of environmental shifts, sensor configuration shifts, and different sets of semantic classes.

\subsection{Implementation Details}
As illustrated in \Cref{fig:LAM_architecture}, the LAM architecture has three fully-connected layers which consist of 32, 64, and 128 channels, respectively. We train LAM on the same train/validation split as the source model. However, for nuScenes~\cite{nuscenes2019} totaling around 400K sweeps, in comparison to SemanticKITTI~\cite{behley2019iccv} ($\sim$20K), SemanticUSL~\cite{jiang2020lidarnet}($\sim$18K), and SemanticPOSS~\cite{pan2020semanticposs}($\sim$3K), running cross-frame ensembling on the entire set of sequences is costly in terms of both memory and storage. Therefore, when using the nuScenes dataset to train LAM, we subsample the training data while validation is kept unmodified. We randomly select 210 sequences from the nuScenes training set, out of the total 700 sequences. The list of these 210 sequences will be included in the future release of the code for reproducibility purposes.

We use the Pytorch~\cite{NEURIPS2019_9015} library for our training code, and we implement the sparse 3D convolutions with the Spconv~\cite{spconv2022} library. Our cross-frame ensembling aggregates 60 frames with a stride of 3 for efficient but dense coverage of the scene. The student model is trained for 25 epochs using the Adam~\cite{adam} optimizer with a learning rate of $1e\!-3$ and with other optimization hyperparameters set to default.

We pre-compute the neighbors $\NN(\pb)$ as well as the model predictions for label generation, which significantly reduces the pre-processing time during the student model training. We use the Faiss~\cite{faiss} library to run a nearest neighbor search to find the 60 nearest neighbors for each query point. We employ an $\epsilon=0.2m$ radius filtering of found nearest neighbors exclusively in our SemanticKITTI and nuScenes experiments since such radius filtering does not yield any benefits in other experiments. We maintain a fixed size set for all points with zero padding.

We train the source model with 4 types of data augmentation. 
We apply 1) random rotation around the z-axis with an angle sampled from $[-45^\circ, 45^\circ]$, 2) flip augmentation by randomly flipping $x$ and/or $y$ coordinates, 3) random scaling with a value sampled from $[0.95, 1.05]$, and 4) random translation with zero-mean Gaussian noise and a standard deviation of 0.1. We refer to this setting as the \textit{basic augmentation} scheme.

As shown in the ablation study, the student model with self-ensembles performs better when trained with a more aggressive, \textit{intense augmentation} scheme. In this setting, we use the same set of augmentations but increase the range of values. Specifically, we increase the random scale to $[0.9, 1.1]$ and set the standard deviation for random translation to 0.5.

\subsection{Comparisons on SemanticKITTI and nuScenes}\label{sec:KITTI2NUS}

SemanticKITTI and nuScenes datasets focus on urban driving, and hence their semantic labels are specific to the urban roads, including but not limited to road surface, sidewalk, pedestrian, car, etc. For this experiment, we adopt the results from four prior works as the baseline. The baseline methods and our method all use the MinkowskiNet~\cite{choy20194d} architecture to make fair comparisons.

The SemanticKITTI and nuScenes datasets present challenges from significant sensor configuration shifts, collected with a 64-beam HDL-64E and with a 32-beam VLP-32, respectively. Additionally, the sensor from nuScenes is facing the right side of the vehicle, while the sensor from SemanticKITTI is facing forward, resulting in a $-90^\circ$ rotation difference between them around the $z$-axis. From minor differences such as axis rotation, sensor height, and viewpoint angle, to major variances such as beam pattern and resolution, the domain gap in sensor configuration makes these adaptation scenarios highly challenging.

As shown in \Cref{tab:kittinusc}, the source models without any DA experience severe degradation in target domain performance. In stark contrast, our method improves over the source model by 14.1\% mIoU and 10.9\% mIoU in each scenario. We also compare our method with two state-of-the-art methods Complete \& Label~\cite{comp_label}, which uses aggregated LiDAR scans as a dense representation, and Graph Matching~\cite{graph_matching}, which uses graph-based feature extraction to align local features across domains. Our method surpasses both methods by a large margin. 
As we elaborate in \Cref{subsec:ablation_study}, our success stems from our structural point cloud subsampling augmentation, cross-frame ensembling, and LAM.

\begin{table}[!t]\centering
\scriptsize
\begin{tabular}{c|c|c|c}
\toprule
Source & Target & Method & mIoU (\%) $\uparrow$ \\
\drule
\multirow{7}{*}{KITTI} & KITTI & Source & 45.80 \\
\cmidrule{2-4}
& \multirow{6}{*}{NUS} & Source & 27.75 \\
& & SqueezeSegV2$^*$~\cite{squeezesegv2} & 10.10 \\
& & SWD$^*$~\cite{lee2019sliced} & 27.70 \\
& & Complete \& Label~\cite{comp_label} & 31.60 \\
& & Graph Matching~\cite{graph_matching} & 37.30 \\
& & \textbf{LiDAR-UDA} & \textbf{41.84} \\
\toprule
\multirow{6}{*}{NUS} & NUS & Source & 50.72 \\
\cmidrule{2-4}
& \multirow{5}{*}{KITTI} & Source & 23.17 \\
& & SqueezeSegV2$^*$~\cite{squeezesegv2} & 13.40 \\
& & SWD$^*$~\cite{lee2019sliced} & 24.50 \\
& & Complete \& Label~\cite{comp_label} & 33.70 \\
& & \textbf{LiDAR-UDA} & \textbf{34.04} \\
\bottomrule
\end{tabular}
\vspace{0.2cm}
\caption{Comparison of methods for SemanticKITTI (KITTI) and nuScenes (NUS) datasets. MinkowskiNet~\cite{choy20194d} architecture is adopted for all methods. $^*$ Results from~\cite{comp_label}}\label{tab:kittinusc}
\end{table}

\begin{table*}[!ht]\centering
\scriptsize
\scalebox{0.93}{
\begin{tabular}{c|c|c|c|c|c|c|c|c|c|c|c|c|c|c|c}
\toprule
Source &Target &Method &Person &Rider &Car &Trunk &Vegetation &Sign &Pole &Object &Building &Fence &Bike &Ground &mIoU \\
\drule
\multirow{15}{*}{KITTI} &\multirow{2}{*}{KITTI} & Source (LiDARNet) & 62.09 & 74.21 & 93.59 & 61.15 & 91.11 & 37.99 & 57.94 & 50.36 & 84.82 & 54.64 & 15.48 & 94.13 & 64.79 \\
& &Source (Ours) & 42.06 & 69.11 & {94.89} & {60.53} & {85.58} & {31.86} & {59.00} & {39.94} & {88.50} & {47.58} & {9.49} & {94.34} & {60.24} \\
\cmidrule{2-16}\morecmidrules\cmidrule{2-16}

& \multirow{5}{*}{USL} & Source & 33.90 & 0.00 & 27.45 & 10.68 & 36.89 & 16.20 & 12.72 & 5.68 & 41.61 & 3.55 & \textbf{31.60} & 75.95 & 24.69\\
& & CyCADA & 0.38 & 0.00 & 28.70 & 13.83 & 57.11 & 20.70 & 23.83 & 3.78 & 53.14 & 22.30 & 9.24 & 72.36 & 25.45\\
& & LidarNet & 33.17 & 0.00 & 67.75 & 38.95 & \textbf{85.60} & 49.94 & 43.44 & 8.94 & \textbf{72.86} & \textbf{44.06} & 23.07 & \textbf{93.18} & 46.75 \\
\cmidrule{3-16}
& & {Source} & {42.00} & {0.00} & {69.18} & {35.08} & {82.94} & {6.80} & {43.41} & {13.23} & {68.02} & {42.75} & {2.13} & {92.45} & {41.51} \\
& & \textbf{LiDAR-UDA} & \textbf{49.50} & 0.00 & \textbf{78.42} & \textbf{53.78} & 85.47 & \textbf{58.56} & \textbf{59.97} & \textbf{19.42} & 69.70 & 32.16 & 0.00 & 92.70 & \textbf{49.97} \\
\cmidrule{2-16}\morecmidrules\cmidrule{2-16}

&\multirow{5}{*}{POSS} & Source & 22.77 & 1.78 & 35.91 & 16.86 & 39.84 & 7.08 & 9.73 & 0.18 & 57.03 & 1.64 & \textbf{18.17} & 41.99 & 21.08 \\
& & CyCADA & 0.00 & 0.00 & 0.00 & 1.45 & 0.00 & 0.00 & 0.00 & 0.00 & 0.00 & 0.00 & 0.00 & 0.00 & 0.12\\
& &LidarNet & 31.39 & \textbf{23.98} & \textbf{70.78} &21.43 &60.68 & \textbf{9.59} &17.48 & \textbf{4.97} & \textbf{79.53} &12.57 &0.78 & \textbf{82.41} &34.63 \\
\cmidrule{3-16}
& & Source & 31.76 & 9.07 & 46.81 & 22.69 & 60.61 & 0.05 & 26.51 & 2.51 & 70.87 & 23.3 & 1.05 & 75.06 & 30.86 \\
& & \textbf{LiDAR-UDA} & \textbf{65.59} & 2.19 & 64.12 & \textbf{27.49} & \textbf{65.40} & 6.44 & \textbf{36.57} & 4.19 & 75.21 & \textbf{40.31} & 0.00 & 75.06 & \textbf{38.55} \\
\toprule
\multirow{13}{*}{POSS} & \multirow{2}{*}{POSS} & Source (LiDARNet) & 64.47 & 48.25 & 85.77 & 29.71 & 62.71 & 27.29 & 38.19 & 8.07 & 84.90 & 48.50 & 65.56 & 72.56 & 53.00\\
& & {Source (Ours)} & {73.65} & {34.18} & {69.48} & {27.58} & {71.11} & {28.24} & {24.43} & {13.90} & {79.71} & {44.11} & {47.73} & {78.93} & {49.42}\\
\cmidrule{2-16}\morecmidrules\cmidrule{2-16}
&\multirow{4}{*}{KITTI} & Source & 5.20 & 0.50 & 22.57 & 0.54 & 44.00 & 1.90 & 12.83 & 0.08 & 43.09 & 0.70 & 0.40 & 5.62 & 11.45 \\
& & LiDARNet & \textbf{23.64} & 24.86 & 71.31 & 23.67 & 72.38 & 4.17 & {31.28} & \textbf{2.48} & 59.41 & 0.36 & 0.53 & 68.68 & 32.06 \\
\cmidrule{3-16}
& & {Source} & {14.17} & \textbf{48.21} & {63.57} & {18.93} & {65.43} & {1.63} & {9.47} & {0.16} & {55.75} & {1.06} & {0.52} & {84.07} & {30.25} \\
& & \textbf{LiDAR-UDA} & 11.65 & 39.1 & \textbf{83.17} & \textbf{27.46} & \textbf{76.95} & \textbf{8.60} & \textbf{33.14} & 0.28 & \textbf{66.43} & \textbf{1.73} & \textbf{1.82} & \textbf{92.95} & \textbf{37.77} \\
\cmidrule{2-16}\morecmidrules\cmidrule{2-16}
&\multirow{4}{*}{USL} & Source & 2.45 & 0.00 & 16.15 & 1.21 & 27.94 & 1.34 & 4.52 & 0.62 & 44.37 & 0.12 & 1.16 & 8.05 & 8.99 \\
& & LiDARNet & 30.38 & 0.00 & 45.73 & {28.69} & 63.08 & \textbf{22.29} & \textbf{33.92} & 4.12 & 63.70 & 1.89 & 9.42 & 77.49 & 31.73 \\
\cmidrule{3-16}
& & Source & 35.07 & 0.00 & {51.59} & 27.51 & 75.84 & 13.36 & 25.98 & 0.05 & 65.47 & 1.10 & \textbf{11.21} & 90.95 & 33.18\\
& & \textbf{LiDAR-UDA} & \textbf{45.49} & {0.00} & \textbf{61.74} & \textbf{31.37} & \textbf{82.67} & {15.30} & {25.19} & \textbf{15.87} & \textbf{68.10} & \textbf{8.20} & {8.19} & \textbf{92.30} & \textbf{37.87}\\

\bottomrule
\end{tabular}
}
\vspace{0.1cm}
\caption{Comparison of methods for SemanticKitti (KITTI), SemanticPOSS (POSS), and SemanticUSL (USL) datasets. MinkowskiNet~\cite{choy20194d} architecture is adopted for LiDAR-UDA, while CyCADA~\cite{cycada} and LiDARNet adopted the LiDARNet architecture~\cite{jiang2020lidarnet}, a boundary-aware variant of SalsaNext~\cite{cortinhal2020salsanext}.}\label{tab:kittipossusl}
\end{table*}

\subsection{Comparisons on SemanticKITTI, SemanticPOSS, and SemanticUSL}
We additionally evaluate our method on adaptation scenarios using SemanticKITTI and SemanticPOSS as the source domains, and SemanticKITTI, SemanticPOSS, and SemanticUSL as the target domains. In contrast to the semanticKITTI/nuScenes scenarios, the corresponding three datasets have relatively similar sensor configurations. Instead, the primary domain gap lies in shifts caused by the different environments since SemanticKITTI is collected strictly from on-road scenarios while SemanticPOSS is collected on the campus area, and SemanticUSL is collected on both campus and off-road testing sites.

We compare our method to LiDARNet~\cite{jiang2020lidarnet}, the current state-of-the-art in UDA for LiDAR segmentation across the SemanticKITTI, SemanticUSL, and SemanticPOSS datasets. LiDARNet adopts the SalsaNext~\cite{cortinhal2020salsanext} architecture backbone and employs adversarial training for adaptation. The SalsaNext backbone utilizes LiDAR intensities in conjunction with the 3D point cloud. However, we have found that the LiDAR intensity domain gap adversely affects the quality of the pseudo labels generated for the target dataset by our source model. Consequently, we choose not to use LiDAR intensities when training the source model and generating pseudo labels for the target set during the first iteration of adaptation. However, we do employ LiDAR intensities in subsequent adaptation steps when training the student model and generating pseudo labels.

As shown in \Cref{tab:kittipossusl}, our method outperforms LiDARNet by 3.2\% mIoU and 3.9\% mIoU in the SemanticKITTI to Semantic USL and SemanticPOSS domain adaptation tasks, respectively. When using SemanticPOSS as the source domain, our method outperforms LiDARNet by 5.7\% mIoU on SemanticKITTI and 6.1\% mIoU on SemanticUSL.

The aforementioned intensity domain gap becomes apparent when comparing the LiDARNet~\cite{jiang2020lidarnet} model, trained with intensity values as input, with our \textit{Source (Ours)} model, trained using the MinkowskiNet architecture without intensity. Despite the SalsaNext model outperforming our MinkowskiNet architecture on the source domains, our source model demonstrates significantly better performance on the target domains, revealing a domain gap in the LiDAR intensities.  Further details can be found in the appendix.

\subsection{Ablation Study}\label{subsec:ablation_study}
In this section, we analyze the individual contributions made by our proposed modules, namely the structural point cloud subsampling, within-frame / cross-frame ensembling, and LAM. Additionally, we explore the applicability of LiDAR-UDA to other point cloud segmentation architectures and self-training strategies.

\textbf{Structural Point Cloud Subsampling} In the upper section of \Cref{tab:subsample_comparison}, we compare the nuScenes pseudo labels obtained from source models trained on SemanticKITTI using various subsampling methods. With a target-to-source ratio of $r=0.5$, our method (Random) drops each row in the range image with a $50\%$ chance. This simulates diverse LiDAR patterns and significantly reduces the domain gap, resulting in an $8.9\%$ improvement in mIoU compared to when no subsampling is applied. On the other hand, regular subsampling, which drops every other row in the LiDAR image, is not as robust to the variability of LiDAR patterns (second row).

\textbf{Cross-frame Ensembling} We also compare the impact of cross-frame ensembling using constant weights (Uniform) and LAM in the lower section of \Cref{tab:subsample_comparison}. The table shows that uniform aggregation boosts the performance by $3.3\%$ in mIoU, and LAM further enhances it by $2.2\%$ over the uniform method. Therefore, \Cref{tab:subsample_comparison} demonstrates that both point cloud subsampling and attention-based cross-frame aggregation are essential for improving the adaptation performance without making specific assumptions on the sensor shift.

\begin{figure}[]
  \centering
  \includegraphics[width=0.9\linewidth]{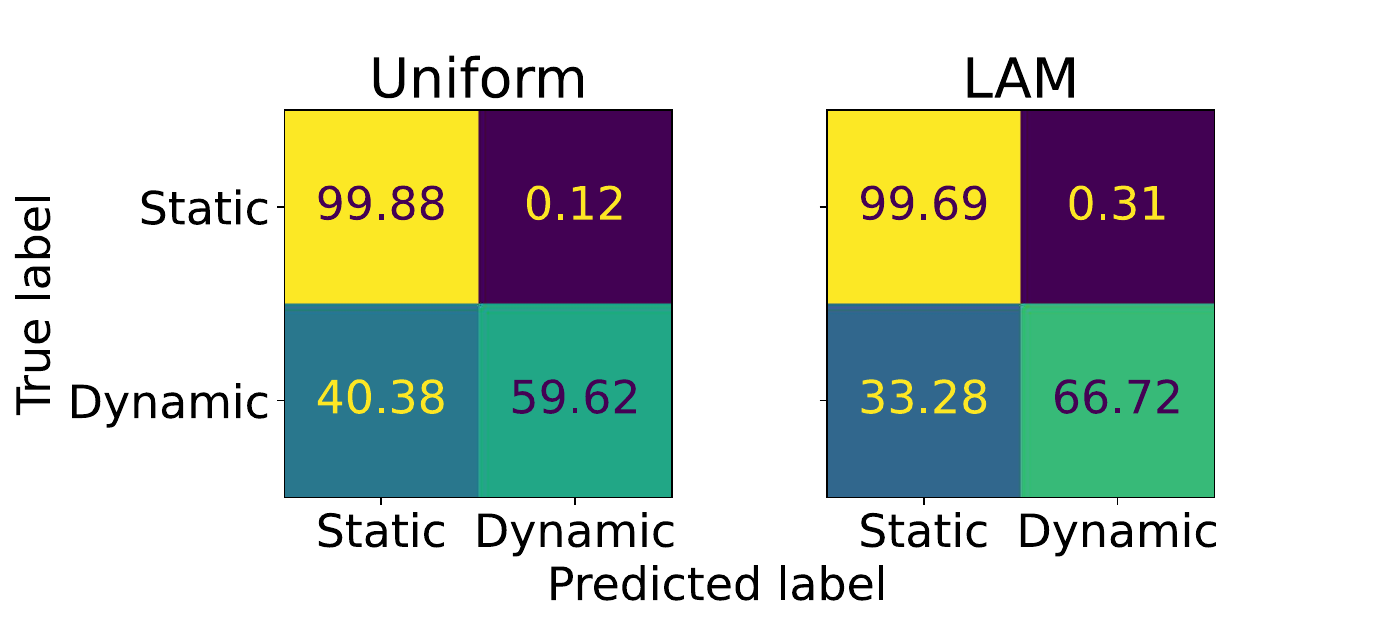}
    \caption{Normalized confusion matrices for static and dynamic classes in SemanticKITTI $\rightarrow$ nuScenes DA experiment. The original row-normalized 10x10 histogram matrix is condensed into a 2x2 matrix by grouping the static and dynamic classes. The results show that LAM outperforms standard uniform weights (Uniform) in predicting dynamic objects.}
    \label{fig:conf}
\end{figure}

To further demonstrate the advantage of LAM, we show the confusion matrices of Uniform and LAM in~\Cref{fig:conf}, where we compare their performance on static (road, terrain, trunk, etc.) and dynamic (car, pedestrian, bicycle, etc.) semantic label classes. We notice that both models perform similarly on static objects, but LAM has significantly fewer false negatives on dynamic objects (\textit{i.e.,} dynamic objects predicted as one of the static classes) than Uniform. This could be explained by the fact that static objects have a higher density in the aggregated point cloud, and uniform weight assignment is potentially biased towards the objects with higher density.

\begin{table}[!t]\centering
\scriptsize
\begin{tabular}{c|c|c|c}
\toprule
Method & Source Subsampling & Cross-frame Ensembling & mIoU (\%) $\uparrow$\\
\drule
1 & \ding{55} & \ding{55} & 27.75 \\
2 & Regular & \ding{55} & 34.97 \\
3 & Random & \ding{55} & 36.64 \\
\midrule
4 & Random & Uniform & 39.96 \\
5 & Random & LAM & \textbf{42.10} \\

\bottomrule
\end{tabular}
\vspace{0.2cm}
\caption{Ablation study for SemanticKITTI $\rightarrow$ nuScenes adaptation. Note that cross-frame ensembling methods in the second part report the \textbf{teacher model} performance on the target domain dataset, and not the student model trained using the teacher as ground truth, which are the final results shown on \Cref{tab:kittinusc}.
}\label{tab:subsample_comparison}
    \vspace{-0.5cm}
\end{table}

\textbf{Within-frame Ensembling} We examine the effect of using the original point cloud and different numbers of subsampled point clouds in within-frame ensembles in \Cref{tab:downsample_comparison}. We observe consistent improvement using a larger number of ensembles, which effectively bridges the gap between different beam patterns. Additionally, comparing the second and third rows reveals the importance of having the original point cloud in the ensemble. We used two random samples alongside the original point cloud in \Cref{sec:KITTI2NUS} to balance time complexity and performance.

\begin{table}\centering
\scriptsize
\begin{tabular}{c|c}
\toprule
Method & mIoU (\%) $\uparrow$ \\
\drule
input & 23.17 \\
2 x random & 23.39 \\
input + 2 x random & 25.50 \\
input + 4 x random & 26.13 \\
input + 8 x random & 26.59 \\
input + 16 x random & 26.84 \\
\bottomrule
\end{tabular}
\vspace{0.2cm}
\caption{Comparison of the within-frame ensembling on the target SemanticKITTI domain for the source model trained on the nuScenes dataset. No adaptation is applied to the model for this comparison. We use MinkowskiNet~\cite{choy20194d} architecture for all methods.}\label{tab:downsample_comparison}
\end{table}

\begin{table}\centering
\scriptsize
\begin{tabular}{c|c|c}
\toprule
&Method & mIoU (\%) $\uparrow$\\
\drule
\multirow{2}{*}{Effects of LAM} & Single-scan + Intense Aug. & 37.52 \\
&LAM + Intense Aug. (Ours in \Cref{tab:kittinusc})& 41.84  \\
\midrule
\multirow{4}{*}{\shortstack{Effects of\\Augmentation}} & Single-scan + Basic Aug. & 37.36 \\
& Single-scan + Intense Aug. & 37.52 \\
& LAM + Basic Aug. & 40.61 \\
& LAM + Intense Aug. (Ours in \Cref{tab:kittinusc}) & 41.84 \\
\bottomrule
\end{tabular}
\vspace{0.2cm}
\caption{Comparision of student models for SemanticKITTI $\rightarrow$ nuScenes DA. 
The single-scan method trains the student model directly from source model pseudo labels without any ensembling. The source model used by all the methods in the table is trained with structural point cloud subsampling.}\label{tab:da_comparison}
\end{table}

\textbf{Effects of LAM \& Data Augmentation on Student}
\Cref{tab:da_comparison} compares the student models trained with 1) single scan pseudo labels, i.e., directly adapting to the target domain model with pseudo labels from the source model, 2) basic data augmentation scheme with LAM, and 3) our intense augmentation with LAM. 

Using LAM yields significant gains over using the single scan pseudo labels with an improvement of 4.3\% mIoU. Meanwhile, the performance gains of 1.2\% mIoU also align with the general understanding that applying stronger augmentation on the student model in a self-training or semi-supervised training framework is beneficial~\cite{araslanov2021self, simclr, simclrv2}. Lastly, we also observe that the basic data augmentation scheme reduces performance in the single scan pseudo label scenario. 

\textbf{Applicability to Other Architectures} To demonstrate the model-agnostic nature of our proposed framework, we test integrating Cylinder3D~\cite{zhu2020cylindrical} into LiDAR-UDA. Cylinder3D utilizes asymmetric cylindrical 3D convolutions, resulting in superior performance compared to MinkowskiNet~\cite{choy20194d}. While maintaining the experimental setup outlined in \Cref{sec:KITTI2NUS}, we opt to use constant weights (Uniform) for this experiment, avoiding the time taken to train LAM. The results in \Cref{tab:kittinusc_cy3d} demonstrate significant improvements for LiDAR-UDA over the source model: 16.1\% mIoU improvement for SemanticKITTI $\rightarrow$ nuScenes and 14.5\% mIoU improvement for nuScenes $\rightarrow$ SemanticKITTI. Furthermore, when compared to the results in \Cref{tab:kittinusc}, using Cylinder3D yields significant improvements, including over our own method using MinkowskiNet.

\begin{table}[!t]\centering
\scriptsize
\begin{tabular}{c|c|c|c}
\toprule
Source & Target & Method & mIoU (\%) $\uparrow$ \\
\drule
\multirow{3}{*}{KITTI} & KITTI & Source & 61.62 \\
\cmidrule{2-4}
& \multirow{2}{*}{NUS} & Source & 32.72 \\
& & \textbf{LiDAR-UDA} & \textbf{48.79} \\
\toprule
\multirow{3}{*}{NUS} & NUS & Source & 74.70 \\
\cmidrule{2-4}
& \multirow{2}{*}{KITTI} & Source & 32.05  \\
& & \textbf{LiDAR-UDA} & \textbf{46.58} \\
\bottomrule
\end{tabular}
\vspace{0.2cm}
\caption{Comparison of methods for SemanticKITTI (KITTI) and nuScenes (NUS) datasets. Cylinder3D~\cite{zhu2020cylindrical} architecture is adopted for all methods.}\label{tab:kittinusc_cy3d}
\end{table}

\begin{figure*}[htb!]
  \centering
  \includegraphics[width=1.0\linewidth]{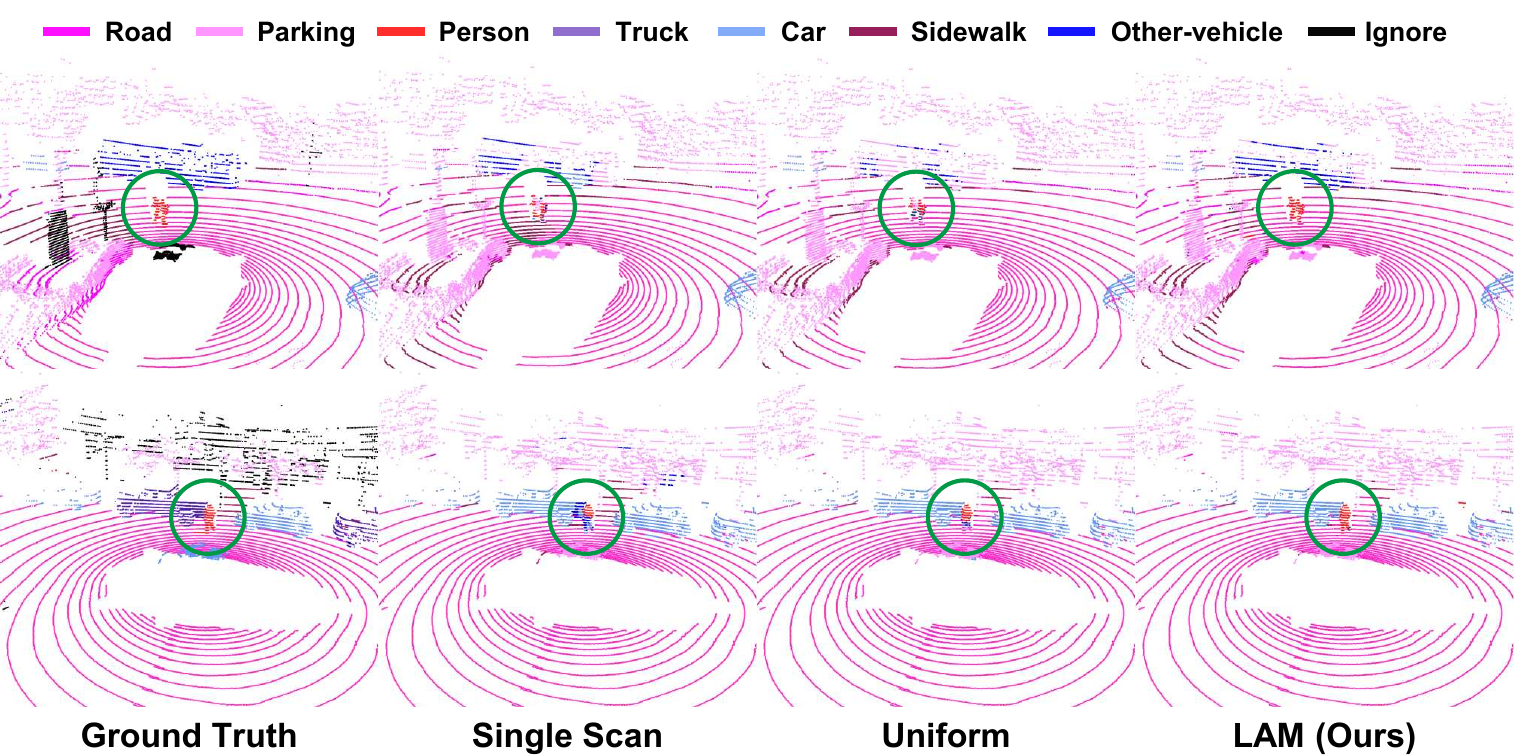}
    \caption{Visualization of two example frames from the held-out target domain data for SemanticKitti $\rightarrow$ nuScenes adaptation scenario. We compare the ground truth against pseudo labels from the base model (single scan), the cross-frame ensembling using uniform weights, and LAM. We circle the specific points of interest, where we see a noticeable improvement in segmenting small objects or sparse parts of a scene with LAM compared to other methods. Note that the unlabeled points are colored in black in the ground truth.
    }
    \label{fig:qualitative}
\end{figure*}

\textbf{LiDAR-UDA with Class-Balanced Self-Training} While our method achieves state-of-art performance in the SemanticKITTI to SemanticPOSS in the adaptation experiment, we observe a reduction of rider and bike classes IoU. We hypothesize that this drop in performance is due to class imbalance, as the rider and bike classes account for only approximately 0.5\% and 5\% of the entire dataset, respectively. To test this hypothesis, we employ CBST~\cite{Zou2018UnsupervisedDA}, a class-balanced self-training framework that avoids the dominance of large classes in pseudo label generation by performing class-wise confidence normalization and selecting a portion of pseudo labels with higher confidence. 

\Cref{tab:cbst_comparison} demonstrates that using CBST with LAM improves the IoU for minor classes, such as rider, traffic sign, pole, and bike. However, this improvement also results in a slight decrease in the overall mIoU. The optimal performance is achieved by setting the pseudo label portion to be $p=20\%$. Please refer to Algorithm 2 of CBST~\cite{Zou2018UnsupervisedDA} for further details on the implemented CBST framework.

Our ensembling and LiDAR augmentation techniques are applicable to various self-training strategies. Thus, exploring different self-training and class-balancing approaches offers promising avenues for future research in LiDAR unsupervised domain adaptation.

\begin{table}[ht!]\centering
\scriptsize
\scalebox{0.9}{
\begin{tabular}{c|c|c|c|c}
\toprule
Classes & Source & LiDAR-UDA &+ CBST ($p=0.2$) & + CBST ($p=0.5$) \\
\drule
Person & 31.76 & \textbf{65.59} & 40.01 & 29.28\\
Rider & 9.07 & 2.19 & \textbf{19.89} & 11.72\\
Car & 46.81 & \textbf{64.12} & 51.55 & 60.84 \\
Trunk & 22.69 & \textbf{27.49} & 25.73 & 27.41\\
Vegetation & 60.61 & \textbf{65.40} & 59.77 & 63.75\\
Traffic-sign & 0.05 & 6.44 & \textbf{21.07} & 4.47\\
Pole & 26.51 & 36.57 & 37.31 & \textbf{39.22}\\
Object & 2.51 & \textbf{4.19} & 1.31 & 1.95\\
Building & 70.87 & \textbf{75.21} & 64.58 & 72.90\\
Fence & 23.30 & \textbf{40.31} & 28.17 & 31.50\\
Bike & 1.05 & 0.00 & \textbf{10.94} & 0.27\\
Ground & 75.06 & 75.06 & 75.73  & \textbf{77.34}\\
\midrule
mIoU & 30.86 & \textbf{38.55} & 36.34 & 35.05 \\
\bottomrule
\end{tabular}
}
\vspace{-0.1cm}
\caption{Comparison of source, Lidar-UDA, and Lidar-UDA + CBST methods on the SemanticKITTI to SemanticPOSS adaptation scenario. CBST denotes class-balanced self-training from Zou \textit{et al.}~\cite{Zou2018UnsupervisedDA}.}\label{tab:cbst_comparison}
\end{table}

\subsection{Qualitative Evaluation}
\Cref{fig:qualitative} shows the effectiveness of our method with a set of examples comparing the base model prediction, uniform weight aggregation, and our method with LAM. We circle the areas of noticeable improvement, where we are able to observe that LAM correctly aggregates model predictions with regard to the geometric and temporal information of the points to segment pedestrians, which are either missed by the base model or washed out when the predicted semantic classes are uniformly aggregated.

\section{Conclusion}
We present LiDAR-UDA, a novel unsupervised domain adaptation framework for LiDAR segmentation. Based on self-training, the framework enables the transfer of model knowledge from the labeled source domain to the unlabeled target domain. Using structural LiDAR point cloud subsampling that reduces the geometric structural gap between the source and target domain input, and cross-frame ensembling that regularizes the self-training, LiDAR-UDA offers an efficient, model-agnostic adaptation method. We demonstrate the effectiveness of our method by surpassing the current state-of-the-art UDA methods on various publicly available LiDAR segmentation datasets. We hope this paper lays a foundation for further exploration of self-training methods for domain adaptation in LiDAR perception.

\newpage
{\small
\bibliographystyle{ieee_fullname}
\bibliography{egbib}
}

\end{document}


\title{LiDAR-UDA: Self-ensembling Through Time for Unsupervised LiDAR Domain Adaptation  \\ \\ {\bf Supplementary Materials}}

\author{First Author\\
Institution1\\
Institution1 address\\
{\tt\small firstauthor@i1.org}
\and
Second Author\\
Institution2\\
First line of institution2 address\\
{\tt\small secondauthor@i2.org}
}
\author{Amirreza Shaban\thanks{Equal Contribution.}\quad
 JoonHo Lee$^*$\quad
 Sanghun Jung$^*$\quad
 Xiangyun Meng\quad
 Byron Boots\quad
 \and
 University of Washington
 }

\maketitle

\section{LiDAR Intensity}
\vspace{-0.2cm}
\begin{table}[!htb]\centering
\scriptsize
\begin{tabular}{c|c|c|c|c}
\toprule
Source & Target (no DA) & source model & Intensity Used & mIoU (\%) $\uparrow$ \\
\drule
\multirow{10}{*}{KITTI} & \multirow{5}{*}{KITTI} & \multirow{2}{*}{MinkowskiNet} & \checkmark & 66.72 \\
 & &  & \ding{55}  & 60.24 \\
\cmidrule{3-5} & & \multirow{2}{*}{SalsaNext} & \checkmark & 58.49 \\
& &  &\ding{55} & 55.81 \\
\cmidrule{2-5}\morecmidrules\cmidrule{2-5} & \multirow{5}{*}{USL} & \multirow{2}{*}{MinkowskiNet} & \checkmark & 38.56 \\
& &  &\ding{55} & 41.51 \\
\cmidrule{3-5} & & \multirow{2}{*}{SalsaNext} & \checkmark & 23.62 \\
& &  &\ding{55} & 36.95 \\
\bottomrule
\end{tabular}
\vspace{0.1cm}
\caption{Generalization performance of MinkowskiNet and SalsaNext trained with and without intensity values as input. We train the source models on SemanticKITTI (KITTI) and evaluate their performance on SemanticUSL (USL). Dropping intensity leads to significantly improved generalization performance on the target domain.}\label{tab:intensity_table}
\end{table}

\vspace{-0.2cm}
Although LiDAR intensity provides additional information for distinguishing geometrically similar objects, we find each LiDAR sensor has significantly different intensity ranges and distribution from others. 
While prior works apply various techniques to utilize the intensity as an additional input~\cite{jiang2020lidarnet, squeezesegv2}, we view intensity matching to remain a non-trivial problem on its own.
As we show in Table~\ref{tab:intensity_table}, using intensity results in degraded generalization performance of the source model -  an average drop of 8.1\% mIoU between MinkowskiNet and SalsaNext.

Furthermore, as our framework \textit{re-introduces} intensity of the target domain as an additional input during the student model training, the proposed LiDAR-UDA method is still able to utilize the intensity information of the target domain in a transferable manner.

\vspace{-0.2cm}
\section{Classwise mIoU Analysis on SemanticKITTI and nuScenes}

We present classwise mIoU results of our method and the source model for the SemanticKITTI $\leftrightarrow$ nuScenes DA scenario in ~\Cref{tab:kittinuscfull}. Our method achieves significant improvements over the source model in most classes, indicating its robust generalization capability. For example, in SemanticKITTI$\rightarrow$nuScenes, our method improves over the source model by $36.06\%$ IoU in Pedestrian and $21.21\%$ IoU in Bus classes. In nuScenes$\rightarrow$SemanticKITTI, our method improves by $31.92\%$ IoU in Drivable and $28.21\%$ IoU in Car classes.

\begin{table*}[!htb]\centering
\scriptsize
\scalebox{0.94}{
\begin{tabular}{c|c|c|c|c|c|c|c|c|c|c|c|c|c}
\toprule
Source & Target & Method & Drivable & Sidewalk & Terrain & Pedestrian & Vegetation & Bicycle & Bus & Car & Motorcycle & Truck & mIoU \\
\drule
\multirow{5}{*}{KITTI} & KITTI & Source & 87.66 & 68.75 & 65.17 & 18.63 & 90.12 & 0.31 & 15.53 & 90.17 & 3.70 & 17.99 & 45.80 \\
\cmidrule{2-14}
& \multirow{6}{*}{NUS} & Source & 79.19 & 32.22 & 21.72 & 4.72 & 73.97 & 0.07 & 4.46 & 55.32 & 3.35 & 2.46 & 27.75 \\
& & SqueezeSegV2$^*$~\cite{squeezesegv2} & - & - & - & - & - & - & - & - & - & - & 10.10 \\
& & SWD$^*$~\cite{lee2019sliced} & - & - & - & - & - & - & - & - & - & - & 27.70 \\
& &  Complete \& Label~\cite{comp_label} & - & - & - & - & - & - & - & - & - & - & 31.60 \\
& & Graph Matching~\cite{graph_matching} & - & - & - & - & - & - & - & - & - & - & 37.30 \\

& & \textbf{LiDAR-UDA (Ours)} & \textbf{87.44} & \textbf{42.31} & \textbf{47.88} & \textbf{40.78} & \textbf{83.22} & \textbf{0.85} & \textbf{25.67} & \textbf{73.48} & \textbf{15.86} & 0.90 & \textbf{41.84} \\
\toprule
\multirow{4}{*}{NUS} & NUS & Source & 91.44 & 52.3 & 58.33 & 48.05 & 86.42 & 3.00 & 24.23 & 80.39 & 28.88 & 34.18 & 50.72 \\
\cmidrule{2-14}
& \multirow{6}{*}{KITTI} & Source & 33.77 & 2.81 & 30.05 & 12.56 & 80.94 & 0.45 & 4.95  & 57.98 & 4.40 & 3.82 & 23.17 \\
& & SqueezeSegV2$^*$~\cite{squeezesegv2} & - & - & - & - & - & - & - & - & - & - & 13.40 \\
& & SWD$^*$~\cite{lee2019sliced} & - & - & - & - & - & - & - & - & - & - & 24.50 \\
& & Complete \& Label~\cite{comp_label} & - & - & - & - & - & - & - & - & - & - & 33.70 \\
& & \textbf{LiDAR-UDA (Ours)} &\textbf{65.69}&\textbf{6.07}&\textbf{54.05}&\textbf{16.49}&\textbf{85.65}&0.00&3.15&\textbf{86.19}&\textbf{13.87}&\textbf{9.30}& \textbf{34.04} \\
\bottomrule
\end{tabular}
}
\caption{Classwise mIoU (\%) of methods on KITTI to Nuscnes and nuScenes to KITTI adaptation scenarios.}\label{tab:kittinuscfull}
\end{table*}
\vspace{-0.2cm}
\section{LAM Analysis: Weight Distributions}

\begin{figure*}
  \centering
  \includegraphics[width=0.7\linewidth]{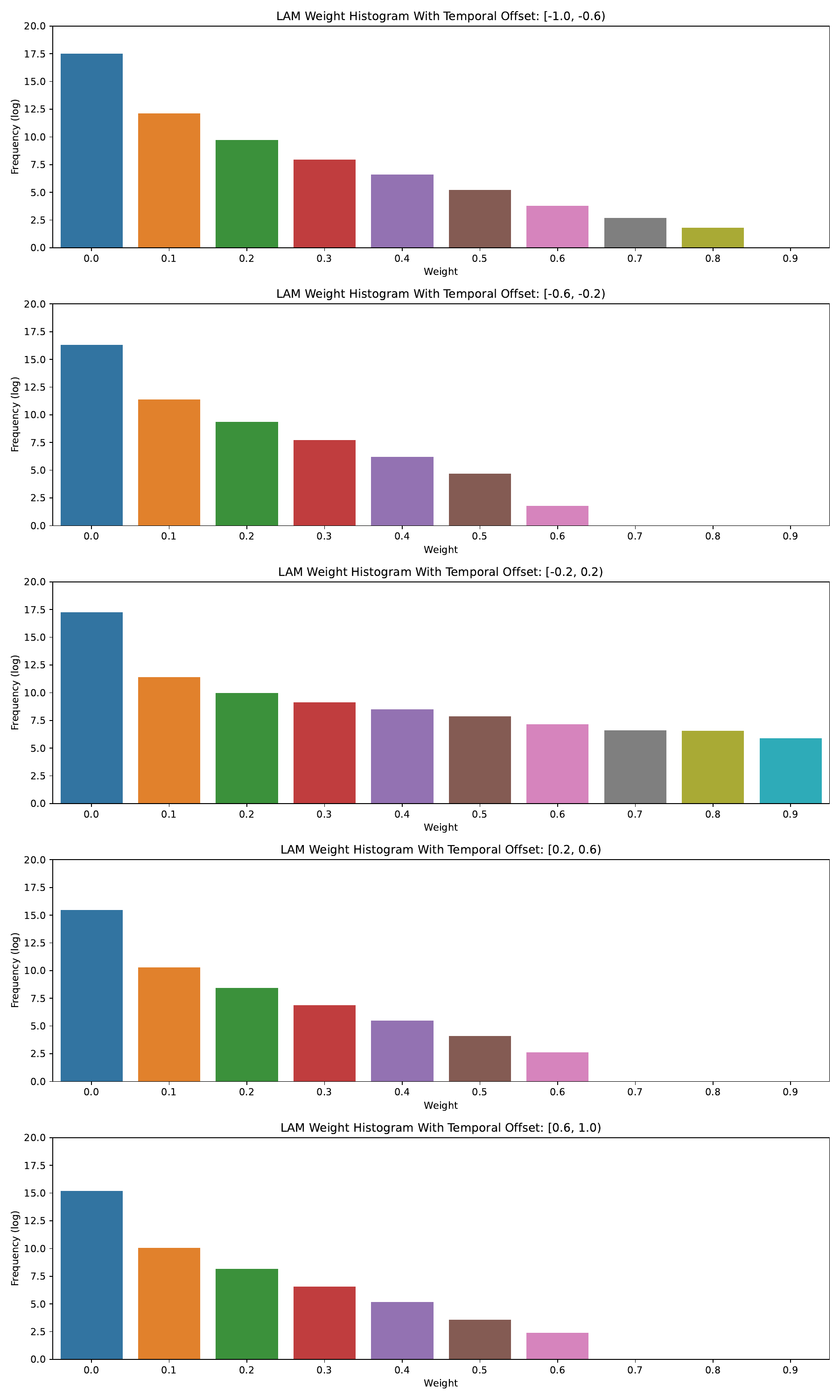}
    \caption{Histogram visualization of LAM weights applied to $\BB_{\epsilon}$ neighbors based on their temporal distance of current frame $t$ normalized between $(-1, +1)$.}
    \label{fig:histogram3}
\end{figure*}

\begin{figure*}
  \centering
  \includegraphics[width=0.7\linewidth]{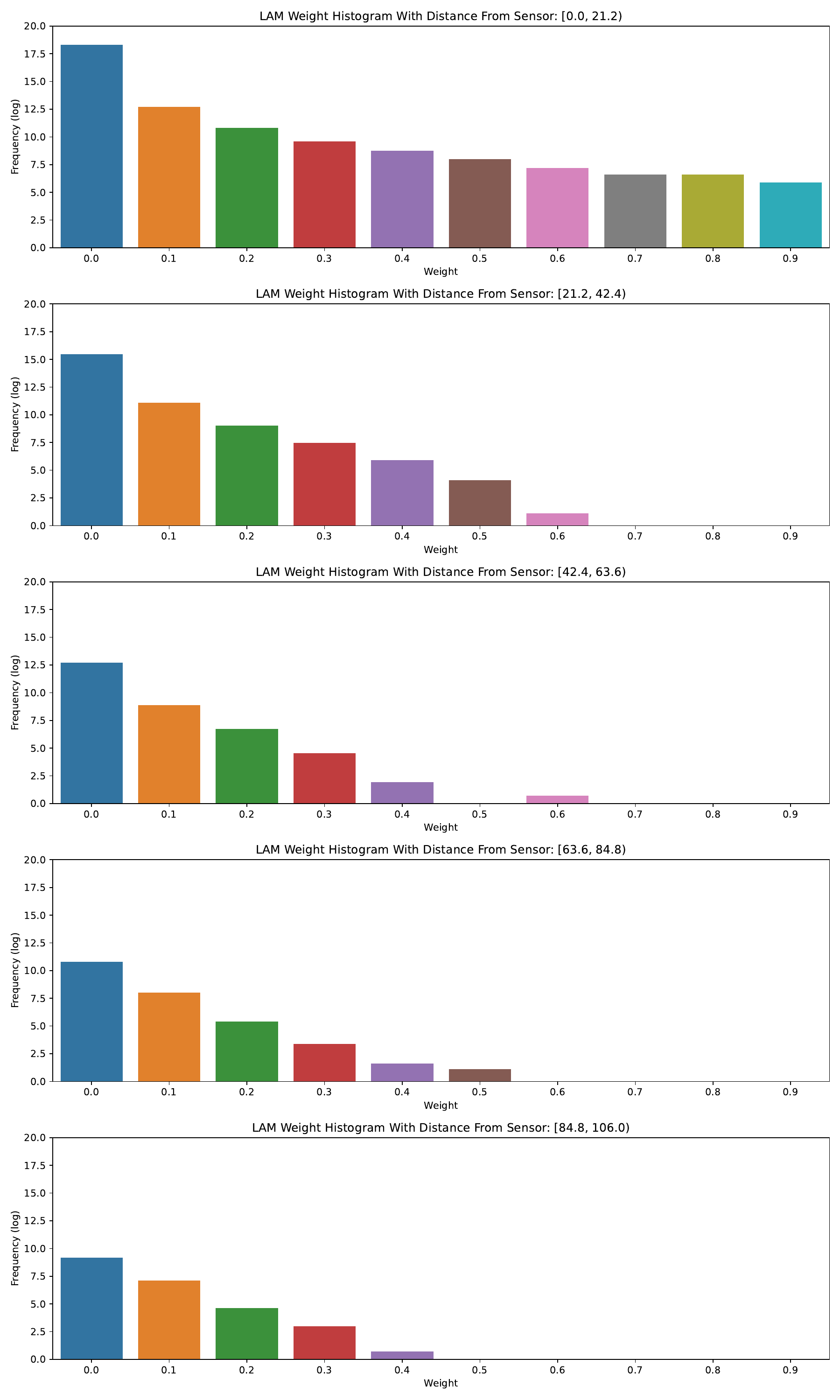}
    \caption{Histogram visualization of LAM weights applied to $\BB_{\epsilon}$ neighbors based on their distance to the sensor location in their corresponding LiDAR scan.}
    \label{fig:histogram2}
\end{figure*}

\begin{figure*}
  \centering
  \includegraphics[width=0.7\linewidth]{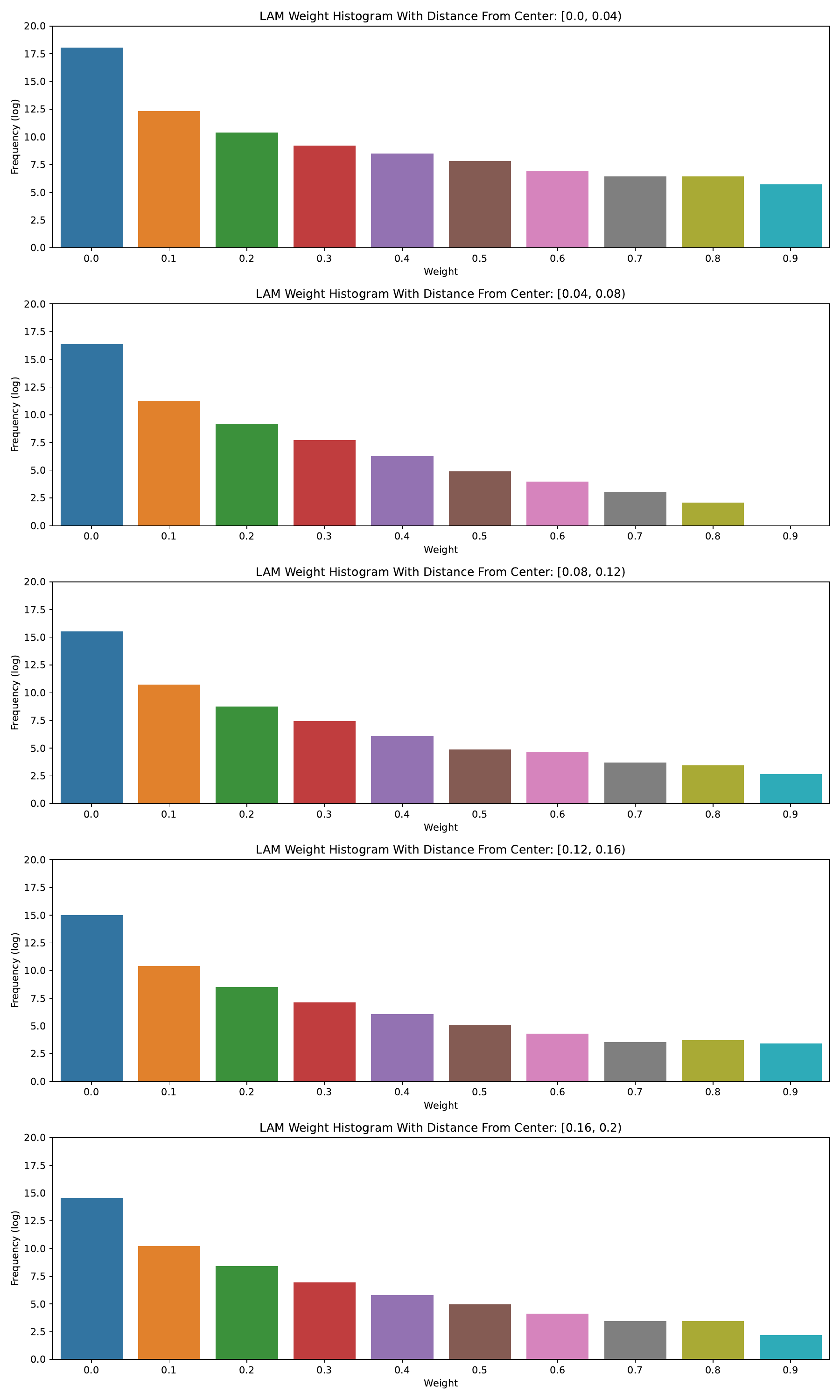}
    \caption{Histogram visualization of LAM weights applied to $\BB_{\epsilon}$ neighbors based on their distance.}
    \label{fig:histogram1}
\end{figure*}

To gain better insight of the weighting dynamics behind LAM, we collected statistics of the weights predicted by LAM in the SemanticKITTI$\rightarrow$nuScenes experiment on various slices of the target dataset, visualized as histograms in \Cref{fig:histogram3,fig:histogram2,fig:histogram1}.

\begin{itemize}[noitemsep]
    \item {\bf Temporal offset:} \Cref{fig:histogram3} shows weight histograms for different temporal distances from the current frame $t$. We find that points with a shorter temporal distance to $t$ (3rd histogram) have relatively higher weights.
    \item {\bf Distance from Sensor:} \Cref{fig:histogram2} shows that LAM favors points closer to the sensor origin in their corresponding LiDAR scan (first row), giving them higher weights. This matches the intuition because the model’s accuracy  typically decreases for objects farther away from the sensor due to the sparsity of the LiDAR point cloud. To illustrate the significance of this feature, we point out that the aggregated point cloud combines $36$ seconds of LiDAR data so if the vehicle speed is as low as $10$ miles per hour, the distance variation of points within an $\epsilon$-ball to their sensor location can exceed $150$ meters.
    \item {\bf Distance from center:} \Cref{fig:histogram1} reports on the distance from the center of $\epsilon$-ball ($||\pb - \pb^\prime||_2$), which shows only a slight decay in the assigned weights as the distance increases, suggesting that LAM depends more on other factors.
\end{itemize}
\vspace{-0.2cm}
Our analysis provides a limited view of predicted weights within each slice, and we overall observe that LAM does not solely rely on only one of the input features above but the combination of all input features to make predictions.

\clearpage
{\small
\bibliographystyle{ieee_fullname}
\bibliography{egbib}
}